\pdfoutput=1

\documentclass[11pt]{article}

\usepackage{EMNLP2023}

\usepackage{times}
\usepackage{courier}
\usepackage{latexsym}
\usepackage{amssymb}
\usepackage{pifont}

\usepackage[T1]{fontenc}

\usepackage[utf8]{inputenc}

\usepackage{microtype}

\usepackage{inconsolata}

\usepackage[autostyle]{csquotes}
\usepackage{graphicx}
\usepackage{makecell}

\usepackage{xcolor}
\definecolor{avocadogreen}{RGB}{86, 130, 3}
\definecolor{seablue}{RGB}{0, 119, 190}
\definecolor{darkyellow}{RGB}{128, 128, 0}

\usepackage{booktabs}
\usepackage{multirow}

\title{Mini-Giants: “Small” Language Models and Open Source Win-Win}

\author{
   Zhengping Zhou \\
  \texttt{zpzhou@cs.stanford.edu} \\
  \And
  Lezhi Li \\
  \texttt{lli2@gsd.harvard.edu} \\
  \AND
  Xinxi Chen \\
  \texttt{xc336@cornell.edu} \\
   \And
  Andy Li \\
   \texttt{andy@RLAI.institute}  
}

\begin{document}
\maketitle
\begin{abstract}

ChatGPT is phenomenal.
However, it is prohibitively expensive to train and refine such giant models.
Fortunately, small language models are flourishing and becoming more and more competent.
We call them "mini-giants".
We argue that open source community like Kaggle and mini-giants will win-win in many ways, technically, ethically and socially. 
In this article, we present a brief yet rich background,
discuss how to attain small language models, 
present a comparative study of small language models and a brief discussion of evaluation methods,
discuss the application scenarios where small language models are most needed in the real world,
and conclude with discussion and outlook.

\end{abstract}

\begin{figure*}
    \centering
    \includegraphics[width=0.8\linewidth]{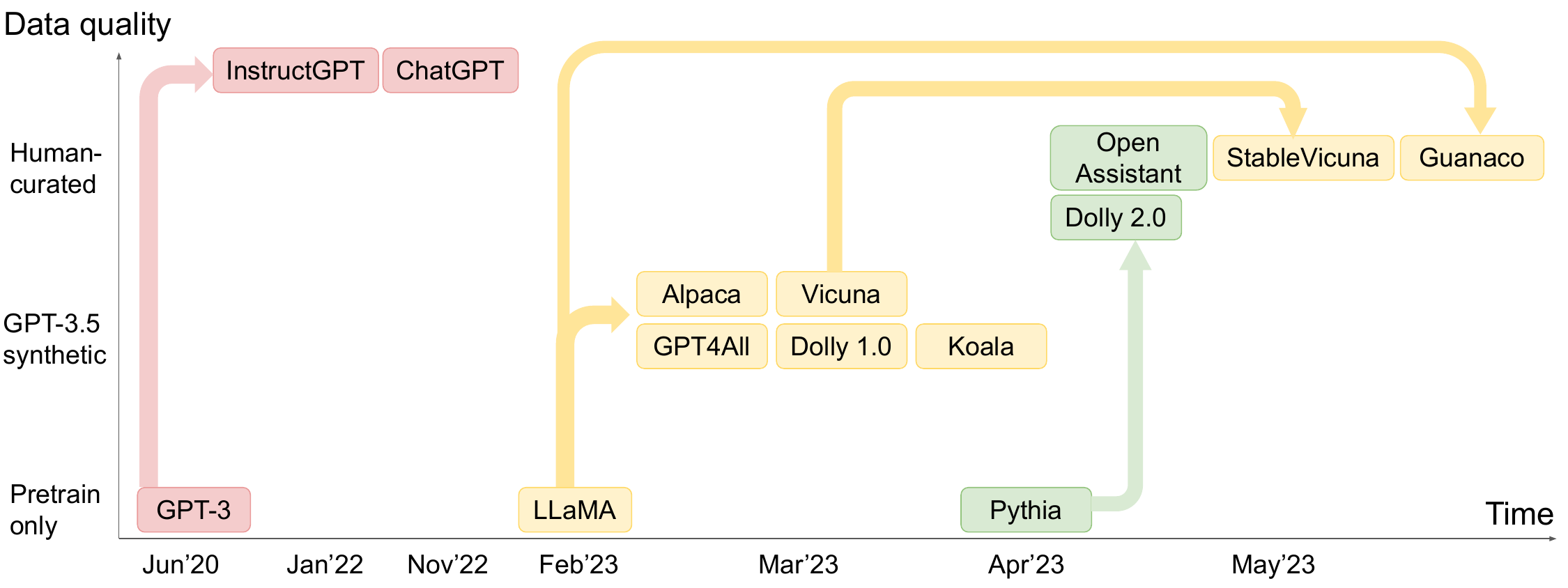}
    \caption{An evolution tree of recently released instruction-following small LMs. The color of the text boxes indicates the openness of the license under which the models are released: red stands for proprietary licenses, yellow stands for non-commercial licenses, and green stands for licenses permissive for commercial use.}
    \label{fig:small_lms_evolution_tree}
\end{figure*}

\section{Introduction}

Large language models (LMs), like ChatGPT and GPT-4, have been taken us by storm. People compare it to the moment of the computer, the moment of the operating system, the moment of the Internet, or the moment of the iPhone. It is considered by many a paradigm shift in NLP and deep learning.

Large language models are large: OpenAI GPT-3 175B parameters, Google PALM 560B, and rumor has it that GPT-4 is as large as 8 × 220B. For most small/medium companies and independent researchers, it is prohibitively expensive to train or update such giant models. In addition, huge consumption of energy for language model training poses a serious concern to the environmental sustainability~\citep{Verdecchia2023GreenAI}.

Recent studies show that network size is not the sole determinant of model performance~\citep{Hoffmann2022Chinchilla}. And thanks to the efforts from the ML open source community as well as private AI companies, we've recently seen more and more "small" LMs created out of these larger models. With their network parameter sizes of around or below 10B, and performance comparable or better than ChatGPT / GPT-4, these "small" LMs are indeed "mini-giants".

In this article, we survey the state-of-the-art for these small language models. We show that compared to their large counterparts, small language/foundation models offer particularly promising opportunities for various industries (including open source ML research and Kaggle competitions) to not only utilize but also to actively participate in the creation/adaptation of modern language models and AI in general. We center our arguments around the 3 key advantages of small models: adaptability, controllability, and affordability.

First of all, smaller models offer better adaptability by being more manageable to modify and fine-tune. In Section \ref{sec:methods}, we present various strategies of creating these small models through optimized fine-tuning techniques. This is important because in most industries (or even in a Kaggle competition), innovation typically arises from the ability to incorporate domain-specific data into the language model or to adjust the model's structure to accommodate their unique requirements. Relying solely on prompt engineering often falls short. Therefore, smaller language models bring forward great benefits to these industries, offering the much-needed flexibility for adaptation, allowing them to full leverage the power of AI and thus catalyzing innovation within them.

Second, smaller models can run on local infrastructure without resorting to GPU-rich third parties, improving the model's controllability by ensuring model users' autonomous data governance and result monitoring. In Section \ref{section:application}, we discuss real world scenarios where small language models fill in the gaps when their large counterparts are unacceptable due to privacy concerns. In Section \ref{sec:survey}, we also look into strategies for customized instruction following and other pioneer research directions for small models, underpinning the relevance of smaller language models in ensuring compliance and mitigating the risk of misinformation. Understanding and managing the way a model operates, the data it accesses, and the outputs it produces, form the cornerstone of responsible AI usage.

Another crucial aspect of the superiority of small language models, is affordability. Taking an average Kaggle competitor as an example. The demanding nature of a Kaggle competition requires the competitor to iterate on the modeling solutions, often times by integrating a variety of data sources and trying different architectures. This necessitates transparent model components and fast iteration pace, which is at odds with the resource requirements that super large language models impose. Having access to fast and inexpensive training / inferencing options, means that he/she will not have to face the trade-off between being constrained in their innovation space, and moving away from language model solutions entirely. As another example which is elaborated in Section \ref{section:application}, privacy-sensitive sectors such as finance and healthcare face a more pressing challenge of choosing between regulation risks and the prohibitive cost of training massive models in-house. Small language models provide the opportunity for them to conform with regulations while not missing out on the power of latest AI technologies.

\subsection*{Outline}

In the following sections, we 
first present a brief yet rich background. Next, we discuss how to attain small foundation models, including parameter reduction and efficient training/fine-tuning techniques. Then we present a comparative study of “small” foundation models a brief discussion of evaluation methods. After that, we discuss the application scenarios where small foundation models are most needed in the real world. We conclude with discussions and an outlook.

\begin{table*}[t]
\centering
\small
\begin{tabular}{ l l r}
\toprule
\textbf{Model} & \textbf{Application} & \textbf{Reference} \\
\midrule

AlphaFold&  Protein folding &\citet{Tunyasuvunakool2021AlphaFold} \\ 
Codex&  Coding & \citet{Chen2021Codex}\\ 
AlphaCode&  Coding & \citet{Li2022AlphaCode}\\ 
RT-1&  Robotics & \citet{Brohan2022RT1}\\ 
BiomedGPT&  Biomedical & \citet{Zhang2023BiomedGPT}\\ 
Clinical Camel&  Clinical & \citet{Toma2023ClinicalCamel}\\ 
BloombergGPT& Finance & \citet{Wu2023BloombergGPT}\\ 
FinGPT& Finance & \citet{Yang2023FinGPT}\\ 
Med-PaLM 2& Medical & \citet{Singhal2023MedPaLM2}\\ 
MusicLM& Music & \citet{Agostinelli2023MusicLM}\\ 
AudioGPT& Audio & \citet{Huang2023AudioGPT}\\ 
\bottomrule
\end{tabular}
\caption{A (small) sample of specialized LMs} 
\label{SpecializedLMs}
\end{table*}

\section{A brief yet rich background}

\paragraph{The Giants are fast}

ChatGPT set a record for fastest-growing user base: one million users in five days, and 100 million monthly active users in January 2023, two months after launching.   

\citet{Radford2018} introduce generative pre-training for LMs, which could be regarded as \enquote{GPT-1}. 
\citet{Radford2019} introduce GPT-2, an unsupervised multitask learning LM.
\citet{Brown2020} introduce GPT-3, a few-shot learning LM, popularizing the concept of in-context learning.
\citet{ChatGPT} introduces ChatGPT and \citet{OpenAI2023GPT4} introduces GPT-4.

There are many LMs released in recent years:
Google BERT, Bidirectional Encoder Representations from Transformers~\citep{Devlin2019BERT},
Google T5, Text-To-Text Transfer Transformer~\citep{Raffel2020T5},
Google LaMDA, Language Model for Dialogue Applications~\citep{Thoppilan2022LaMDA},
Google PaLM, Pathways Language Model~\citep{Chowdhery2022PaLM},
Deepmind Sparrow~\citep{Glaese2022Sparrow},
Anthropic Claude~\citep{Bai2022Claude},
Deepmind Chinchilla~\citep{Hoffmann2022Chinchilla}
Nivedia Megatron-Turing NLG~\citep{Smith2022MegatronTuringNLG},
Deepmind Gopher~\citep{Rae2022Gopher},
HuggingFace BLOOM~\citep{BigScience2023BLOOM},
and Meta LLaMA, Large Language Model Meta AI~\citep{Touvron2023LLaMA}.


\paragraph{Language models as experts}

Besides general purpose LMs as above, there are many specialized models for various application, e.g., Table~\ref{SpecializedLMs} shows a sample of them.

\paragraph{Language and functional competence}

\citet{Mahowald2023} study language competence vs thought competence of LMs and show impressive but imperfect formal linguistic competence, i.e., \enquote{knowledge of rules and patterns of a given language},
yet failures on many tests requiring functional linguistic competence, i.e.,\enquote{a host of cognitive abilities required for language understanding and use in the real world}. 

Then we can leverage LMs’ competence as a good model of language, e.g., by prompt engineering. We can also manage to improve the functional competence, e.g., factuality, safety, and planning. 
With the capacity of in-context learning~\citep{Brown2020}, prompting is a natural and popular way to utilize LMs.
Prompting is the user interface for LMs, and can be formed with advanced methods like search and coding, e.g., Tree of Thoughts (ToT)~\citep{Yao2023ToT}, AdaPlanner~\citep{Sun2023AdaPlanner}, Code as Policies~\citep{Liang2023Code}.
Fine-tuning can improve LMs further. 
A parameter efficient approach makes fine-tuning large LMs feasible considering the cost~\citep{Hu2021LoRA, Ding2023DeltaTuning, Ruder2022Tutorial}.
Augmenting LMs with tools can achieve various functionalities.

To approach artificial general intelligence (AGI) from language models, \citet{Mahowald2023} suggest that, “instead of or in addition to scaling up the size of the models, more promising solutions will come in the form of modular architectures …, like the human brain, integrate language processing with additional systems that carry out perception, reasoning, and planning”. The authors believe that “a model that succeeds at real-world language use would include -- in addition to the core language component -- a successful problem solver, a grounded experiencer, a situation modeler, a pragmatic reasoner, and a goal setter”. 

\paragraph{Augmented LMs with tools}

A natural way to harnesses the language competence of LMs is by utilizing tools like a search engine, a vector database, a code interpreter, or a solver to handle tasks, e.g., 
LangChain\footnote{\url{https://langchain.com}}, 
HuggingGPT~\citep{Shen2023HuggingGPT}, 
Visual ChatGPT~\citep{Wu2023VisualChatGPT}, 
TaskMatrix.AI~\citep{Liang2023TaskMatrixai}, 
RCI~\citep{Kim2023RCI}, 
LLM+P~\citep{Liu2023LLMP}, 
ChemCrow~\citep{Bran2023ChemCrow},  
etc.
See \citet{Mialon2023Augmented} for a survey about augmented LMs.

Domain expertise is still required, e.g., the ChemCrow~\citet{Bran2023ChemCrow} authors mention that “However, it is important to emphasize that potential risks may arise for non-experts who lack the chemical reasoning to evaluate results or the proper lab training, as conducting experiments still necessitates thorough laboratory experience.” and the director of the movie trailer mentions that “For those who believe that AI will do everything for you: No!” and “I’ll always prefer to put my own heart \& soul in.”~\footnote{\url{https://twitter.com/ChristianF369/status/1651607149804498946}}

\paragraph{Mini-Giants are coming}

Following the leakage of LLaMA~\citep{Touvron2023LLaMA}, 
many “small” LMs appear in the open source community, with neural network parameter sizes of around 10B or smaller, e.g., 
Alpaca~\citep{Taori2023Alpaca}, 
Dolly~\citep{Conover2023Dolly}, 
Koala~\citep{Geng2023Koala}, 
Vicuna~\citep{Chiang2023Vicuna}, 
StableLM~\citep{StableLM2023}, 
ChatGLM~\citep{Du2022GLM, Zeng2023GLM130B}, 
Guanaco~\citep{Dettmers2023QLoRA},
Pythia~\citep{Biderman2023Pythia},
GPT4All\footnote{\url{https://github.com/nomic-ai/gpt4all}},
Open-Assistant\footnote{\url{https://github.com/LAION-AI/Open-Assistant}},
ColossalChat~\citep{You2023ColossalChat}.

See~\citet{Kim2023LLMs} for a list of open sourced fine-tuned LMs. In Section \ref{sec:survey}, we will discuss and compare these mini-giants in details.

\paragraph{Discussions \& debates abound}

There are all sorts of discussions \& debates, e.g. discussions about AI alignment with human value from \citet{Russell2019, Mitchell2020, Christian2021Alignment}. Table~\ref{Debates} lists a few representative examples. 

\begin{table*}[t    ]
\centering
\small
\begin{tabular}{l  r }
\toprule
\textbf{Issue to discuss} & \textbf{Reference} \\
\midrule
The dangers of stochastic parrots& \citet{Bender2021Parrots} \\ 
 Limitation of neural networks& \citet{Deletang2023Chomsky}  \\ 
Limitation of autoregressive models& \citet{Lin2021Limitations} \\ 
 Lack of causality& \citet{Jin2023Causality} \\ 
Lack of compositionality& \citet{Dziri2023Faith}  \\ 
Lack of recursion& \citet{Zhang2023Recursive} \\ 
 Limitation of scaling laws & \citet{Deshpande2023Scale} \\ 
 Limitation of scaling laws & \citet{Mckenzie2023Inverse} \\ 
 Model collapse& \citet{Shumailov2023Curse} \\ 
 Artificial general intelligence (AGI)& \citet{Marcus2023AGI} \\ 
 Evaluation of AI& \citet{Burnell2023Evaluation} \\ 
 Distortion of human beliefs& \citet{Kidd2023Distort} \\ 
 Social norms& \citet{Browning2023SocialNorms} \\ 
 Risks and benefits& \citet{Goldman2023Cho} \\ 
 Existential risk& \citet{Bengio2023Rogue} \\ 
 Court hearing due to hallucination& \citet{Novak2023Court} \\ 
 Risk of further concentration of wealth& \citet{Chiang2023McKinsey} \\ 
 Eight things to know& \citet{Bowman2023Eight} \\ 
\bottomrule
\end{tabular}
\caption{Discussions and debates of LMs} 
\label{Debates}
\end{table*}

\section{How to make large foundation models "small"}
\label{sec:methods}
Since the advent of ultra-capable large foundation models like ChatGPT and StableDiffusion, numerous efforts have been devoted to address the primary challenges for their wide-spread utilization: their humongous parameter sizes and the sheer time and compute resources needed to fine-tune them. Within 2 years, the research and open source community have arrived at several strategies to cope with this issue, which we will discuss in this section. 

We classify these strategies into 2 groups: ones that directly reduce the parameter sizes, and ones that makes fine-tuning large models more efficient.

\subsection{Foundation models with reduced parameters}
\paragraph{Chinchilla} 
    ~\citep{Hoffmann2022Chinchilla} is the first influential study on computational efficiency of modern large language models. It put forward the argument that given a compute budget, the best model is attained not by larger parameter size, but by more training data tokens. Based on this principle, the authors produced the Chinchilla 70B model which out-performs prior large models 4 times as large, with the same amount of compute. 
\paragraph{LLaMa} 
    ~\citep{Touvron2023LLaMA} further reduces the parameters and released a series of models ranging from 7 to 65B parameters, following the Chinchilla computation rule. Notably, the paper used only publicly available datasets as training corpus and proved comparable performance as closed source counterparts. This, as commented by~\citep{Harris2023Canon}, started a revolution of open source LLM models. Along with parameter reduction, another contribution by the authors is efficient implementation of multi-headed attention layers through the open source \emph{xformers} library, which optimizes the memory consumption in training.

\subsection{Efficient fine-tuning strategies for foundation models}
Compared with building even more compact models, the majority of research work by the ML community in the direction of "smaller" foundation models, is around making them easier to fine-tune. Here we list several key strategies to achieve this.
\paragraph{Adapter} 
    ~\citep{houlsby2019parameter} is a strategy to add NN layers after existing layers (usually transformer blocks) in pretrained foundation models, so that they can be adapted to custom tasks without changing the weights of existing layers. This paper proposes an adapter module with two linear layers plus a non-linear activation in between. The first layer projects the hidden state to a lower-dimensional space, and the second layer projects it back to the original dimension. A newer paper~\citep{lin2020exploring} recommended only one linear layer plus an additional LayerNorm, as an Adapter module. Adapter achieves near state-of-the-art performance, while adding only a small amount of parameters per task - on GLUE, the added parameters accounted for 3.6\% of the original model.
    
\paragraph{Prefix fine-tuning} 
    ~\citep{Li2021PrefixTuning} Unlike the Adapter architecture that focuses on modifying model behavior via model params, Prefix fine-tuning seeks to train a few params that are used as input prefixes, for each custom sub task. The authors commented that the method is inspired by prompting: similar to prepending a few sentences before a generation task, Prefix-tuning prepends a sequence of trained vectors to the input - just that the prefix vectors do not have to correspond to any real tokens. Compared to full fine-tuning, prefix-fine tuning achieves comparable or better performance with just 0.1\% added parameters.

\paragraph{LoRA}
    ~\citep{Hu2021LoRA} Marks a substantial progress in parameter efficient fine-tuning. Performance-wise, it is more efficient than previous methods like Adapter and Prefix-finetuning. LoRA proposes that we add a low rank, trainable matrix in parallel to the frozen,  pretrained model weights. The activation will be the sum of these two matrices. Formally:
    $$ h = W_0 x + \Delta W x = W_0 x + BAx $$ 

    where $B$ and $A$ are much "thinner" (i.e. low rank), trainable matrices compared to $W_0$ (the frozen pretrained matrix). The use of low rank matrices reduces trainable parameters to as much as by 10,000 times of the original model, compared to a full fine-tune of GPT-3 175B. The article suggests that LoRA can be used next to any model weights, not just transformer layers. The authors claim that LoRA is superior compared to Adapters in that it doesn't introduce additional inference latency; and it's better than Prefix fine-tuning in that it doesn't reduce the available sequence length like the latter does. Further more, since this architectural modification is orthogonal to the ideas of Adapter and Prefix fine-tuning, LoRA can be used in conjunction with them for even better results.

\paragraph{QLoRA}
    \citep{Dettmers2023QLoRA} As an improvement of LoRA, QLoRA proposes optimization methods via quantized low rank fine tuning. Innovations of QLoRA include a 4-bit data type: NormalFloat4, which optimizes information efficiency for normally distributed data (e.g. weights) based on information theory. Apart from that, the paper uses Paged Optimizers (partial optimizer state stored on CPU rather than GPU) to manage memory spikes, like when processing mini batches with long sequence lengths. Experiment results show that fine-tuning using QLoRA reaches 99.3\% of the performance of ChatGPT, and only requires training for 24 hours on one GPU.

\paragraph{ControlNet}
    ~\citep{Zhang2023ControlNet} is proposed as a method to efficiently fine-tune image generation models (diffusion model) on user-defined tasks. Because image generation models in general have a larger design space in terms of user interaction than language models, we list this method here to inspire the readers to consider more complex scenarios of controlling / customizing large foundation model's outputs.
    
    ControlNet copies weights of the original model to a frozen copy (ike all methods mentioned above). The trainable branch consists of an exact same copy as the frozen copy, as well as two convolution layers called "zero convolutions", both before and after the trainable copy. In the fine-tuning forward path, the activation from the trainable copy will be combined with that of the frozen copy by Zero Convolution. The so-called Zero Convolution is just a 1x1 convolution layer that are initiated with both weights and biases being zeros. The result of using ControlNet shows that in some tasks, ControlNets on a personal computer achieve comparable results as commercial models trained on terabytes of GPU memory and thousands of GPU hours.

\section{A brief survey of “small” instruction-following LMs}
\label{sec:survey}

\begin{table*}[t]
    \centering
    \small
    \begin{tabular}{ccccccccccccccc}
    \toprule
        \multicolumn{3}{c}{\bf Basic info} 
        & \multicolumn{3}{c}{\bf Scale} 
        & \multicolumn{4}{c}{\bf Openness} \\
        \cmidrule(lr){1-3}
        \cmidrule(lr){4-6}
        \cmidrule(lr){7-10} 
        \shortstack{Time \\ MM/YY}
        & \shortstack{Model}
        & \shortstack{Institute}
        & \shortstack{\# parameters} 
        & \shortstack{Training \\ hardware cost}
        & \shortstack{Training \\ data size}
        & L
        & \shortstack{I} 
        & \shortstack{TC} 
        & \shortstack{TD} \\
        \midrule
        06/20
        & GPT-3
        & OpenAI
        & 175B
        & 3.64k PT-days 
        & 300B tokens
        & P
        & P
        & P
        & \checkmark \\
        02/23
        & LLaMA-7B
        & Meta
        & 7B
        & 82k GPU-hours
        & 1.4T tokens
        & NC
        & \checkmark
        & \ding{55} 
        & \checkmark \\
        02/23
        & LLaMA-13B
        & Meta
        & 13B
        & 135k GPU-hours
        & 1.4T tokens
        & NC
        & \checkmark
        & \ding{55} 
        & \checkmark \\
        04/23
        & Pythia-7B
        & Eleuther AI
        & 7B
        & 33.5k GPU-hours
        & 300B tokens
        & C
        & \checkmark
        & \checkmark 
        & \checkmark  \\
        04/23
        & Pythia-12B
        & Eleuther AI
        & 12B
        & 72k GPU-hours
        & 300B tokens
        & C
        & \checkmark
        & \checkmark 
        & \checkmark  \\
    \bottomrule
    \end{tabular}
    \vspace{-5pt}
    \caption{\small Comparison of recent base LMs. In the Openness section, L stands for License, I stands for Inference, TC stands for Training Codes, and TD stands for Training Data. In the License column, P stands for Proprietary, NC stands for Non-Commercial, and C stands for permissive for Commercial use.  }
    \label{tab:base_llms}
    \vspace{-2pt}
\end{table*}

\begin{table*}[t]
    \centering
    \small
    \resizebox{\textwidth}{!}{
    \begin{tabular}{ccccccccccccccc}
    \toprule
        \multicolumn{4}{c}{\bf Basic info} 
        & \multicolumn{2}{c}{\bf Scale} 
        & \multicolumn{4}{c}{\bf Openness} \\
        \cmidrule(lr){1-4}
        \cmidrule(lr){5-6}
        \cmidrule(lr){7-10} 
        \shortstack{Time \\ MM/YY}
        & \shortstack{Model}
        & \shortstack{Institute}
        & \shortstack{Backbone}
        & \shortstack{\# parameters} 
        & \shortstack{Training \\ hardware cost}
        & L
        & \shortstack{I} 
        & \shortstack{TC} 
        & \shortstack{TD} \\
        \midrule
        01/22
        & InstructGPT
        & OpenAI
        & GPT-3
        & 1.3B
        & N/A
        & P
        & P
        & P
        & P\\ 
        11/22
        & ChatGPT
        & OpenAI
        & GPT-3
        & N/A
        & N/A
        & P
        & P
        & P 
        & P\\  
        03/23
        & Alpaca-7B
        & Stanford
        & LLaMA-7B 
        & 7B
        & < \$100
        & NC
        & \checkmark
        & \checkmark 
        & \checkmark \\
        03/23
        & GPT4All-Lora
        & Nomic AI
        & LLaMA-7B 
        & 7B
        & \$100
        & NC
        & \checkmark
        & \checkmark 
        & \checkmark \\
        03/23
        & ChatGLM-6B
        & Tsinghua
        & GLM
        & 6B
        & N/A
        & NC
        & \checkmark
        & \checkmark 
        & \ding{55} \\
        03/23
        & Vicuna-7B/13B
        & LMSYS
        & LLaMA-7B/13B 
        & 7B/13B
        & \$140/\$300
        & NC
        & \checkmark
        & \checkmark 
        & \checkmark  \\
        03/23
        & Dolly-6B
        & Databricks
        & GPT-J-6B 
        & 6B
        & < \$30
        & NC
        & \checkmark
        & \checkmark 
        & \checkmark \\
        04/23
        & OASST-12B
        & LAION AI
        & Pythia-12B
        & 12B
        & N/A
        & C
        & \checkmark
        & \checkmark 
        & \checkmark  \\
        04/23
        & Koala-13B
        & Berkeley
        & LLaMA-13B 
        & 13B
        & < \$100
        & NC
        & \checkmark
        & \checkmark 
        & \checkmark  \\
        04/23
        & Dolly-v2-12B
        & Databricks
        & Pythia-12B 
        & 12B
        & N/A
        & C
        & \checkmark
        & \checkmark 
        & \checkmark \\
        04/23
        & StableVicuna-13B
        & Stability AI
        & Vicuna-13B
        & 13B
        & N/A
        & NC
        & \checkmark
        & \checkmark 
        & \checkmark  \\
        05/23
        & Guanaco-7B/13B
        & UW
        & LLaMA-7B/13B
        & 7B/13B
        & < 12 GPU-hours
        & NC
        & \checkmark
        & \checkmark 
        & \checkmark  \\
    \bottomrule
    \end{tabular}
    }
    \vspace{-5pt}
    \caption{\small Comparison of recent instruction-following small LMs. The abbreviations of the column names follow Table \ref{tab:base_llms}.}
    \label{tab:main_llms}
    \vspace{-2pt}
\end{table*}


Over the past few months, we have seen small LMs flourish. See Figure \ref{fig:small_lms_evolution_tree} for an evolution tree.
This is a very fast progressing field, and it is challenging to even keep ahead with the latest progress.
Quoting \citep{tunguz_tweet}, ``Trying to get ahead in AI these days feels like wrestling a rabid 5,000 lbs hippo covered in baby oil''.

\subsection{Closed-source milestones}

\paragraph{GPT-3} 
    \citep{Brown2020} gained public attention when it was released in 2020. As reported by New York Times, it ``generates tweets, pens poetry, summarizes emails, answers trivia questions, translates languages and even writes its own computer programs'' \citep{nytimes-gpt3}. It shows that decent few-shot performance can be achieved  without gradient update, and the unprecedented model scale (175B parameters) is a key ingredient for success.

\paragraph{InstructGPT} 
    Although GPT-3 is already powerful, \citet{InstructGPT} points out that the model output may not align well with human intent and may contain harmful content. For example, when prompted to generate a story, the LM should generate a story instead of rambling around the prompt itself. This necessitated an extra step called \emph{model alignment}, and the desired model behavior is called \emph{instruction-following}. In InstructGPT, this is achieved by applying the reinforcement learning from human feedback (RLHF) \citep{RLHF} technique on top of a GPT-3 backbone. Despite having 100x less parameters, InstructGPT outperforms the unaligned GPT-3 model in human evaluation, giving rise to the phenomenal success of ChatGPT ten months later.

\paragraph{ChatGPT} 
    \citep{ChatGPT} brings AIGC to the attention of the general public. It uses the same technique as InstructGPT, but extends InstructGPT by incorporating dialogue data into the supervised fine-tuning and the RLHF stage. It acquired 1 million users in just 5 days and revolutionizes the way people interact with modern AIs. As a proprietary product, although the web UI is free, the underlying model can only be accessed via a paid API. 

\subsection{Open-source backbone LMs}

\paragraph{LLaMA} 
    Despite the recent success of GPT-3 and ChatGPT, training and deploying LLMs remain a major challenge to the open source community due to the high training infra cost. For instance, the GPT-3 training is estimated to cost millions of dollars. \citep{Touvron2023LLaMA} propose LLaMA, an open source LLM pretrained with public data available at several sizes. Remarkably, the 13B LLaMA model benefited from large scale pretraining data (1.4T tokens), and outperforms the 175B GPT-3 on most benchmarks. It soon becomes a highly influential milestone in the open source world, serving as a powerful yet lightweight backbone for a wide range of subsequent instruction-following small LMs. The non-commercial bespoke license, under which it is released, limits the usage to research purpose only.

\paragraph{Pythia}
    \citep{Biderman2023Pythia} Published two months later than LLaMA, Pythia releases a suite of 16 LLMs ranging from 70M to 12B parameters. Trained with 300B tokens from the Pile \citep{gao2020pile}, it consumed a similar amount of data as GPT-3 but around four times less than LLaMA (see Table \ref{tab:base_llms} for comparison). Released under the Apache 2.0 license, Pythia is free for commercial use, making it an appealing backbone for many subsequent instruction-following small LMs (e.g. Open Assistant \cite{OpenAssistant}, Dolly 2.0 \cite{Conover2023Dolly}).

\subsection{Small LMs trained with GPT synthetic data} 
\label{sec:small_lm_gpt}

Since the release of LLaMA, open-source instruction fine-tuned small LMs emerge at a rapid speed. Viewing LLaMA as an open-source counterpart of GPT-3, these small LMs can be seen as the open-source counterparts of InstructGPT or ChatGPT. Most of them can be fine-tuned under a feasible budget (the training hardware cost can be capped under several hundred dollars).

A major challenge is to obtain high-quality instruction-following data, a key ingredient in the model alignment stage. 
At an early stage, the open-source community tackles this challenge by using GPT-3.5 \cite{ChatGPT} to synthesize the response of a given prompt. This imposes a non-commercial license on the fine-tuned model. 

\paragraph{Alpaca} 
    \citep{Taori2023Alpaca} is the first newborn in this family.  It fine-tunes LLaMA-7B with 52k instruction-following data generated using the self-instruct method, which leverages GPT-3.5 to synthesize prompt-response pairs from a manually created seed set. According to human evaluation, it achieves similar performance to GPT-3.5 on a small sample data. 

\paragraph{GPT4All}
    \citep{gpt4all} fine-tunes LLaMA-7B with 437k prompt-response pairs. The instructions are collected from the \texttt{unified\_chip2} and Stackoverflow Questions, while the responses are generated by GPT-3.5. The model is fine-tuned using the LoRA \citep{Hu2021LoRA} algorithm. Evaluated using the ground truth perplexity on the Self-Instruct \citep{Wang2022SelfInstruct} human evaluation data, GPT4All stochastically outperforms Alpaca.

\paragraph{Vicuna}
    \citep{Chiang2023Vicuna} fine-tunes LLaMA-13B with 70k user-shared conversations with ChatGPT (from ShareGPT.com). Compared to Alpaca, it accounts for multi-turn conversation in training, and made several optimizations to cut the training cost. Vicuna uses GPT-4 as an automatic chatbot judge, based on which it outperforms LLaMA and Alpaca, while achieving more than 90\% quality of ChatGPT. A more rigorous analysis validating this evaluation approach is later presented in the Guanaco work \citep{Dettmers2023QLoRA}.

\paragraph{Koala}
    \citep{Geng2023Koala} is another instruction fine-tuned LLaMA model, with 13B parameters. It is a concurrent effort with Vicuna, released at a similar time. Like Vicuna, it is fine-tuned on ChatGPT-distilled data, with a focus on the dialogue scenario. In human evaluation, Koala achieves comparable or superior results compared to Alpaca. 

\subsection{Small LMs trained with human-curated data} 
\label{sec:small_lm_human}

\paragraph{Dolly 1.0} 
    \citep{Conover2023Dolly} trains a two-year-old GPT-J-6B backbone using the same data as Alpaca, showcasing that the instruction-following capability does not necessarily require state-of-the-art backbone model as long as the data quality is decent. \textbf{Dolly 2.0}, released one month later, upgrades to the newly released Pythia-12B \cite{Biderman2023Pythia} backbone and is instruction fine-tuned using a newly crowd-sourced dataset, \texttt{databricks-dolly-15k} which contains 15k human-generated prompt-response pairs. Notably, it is the first open-source instruction-following small LM that permits commercial use.  

\paragraph{Open Assistant}
    \citep{OpenAssistant} uses LLaMA-13B and Pythia-12B as the backbones, allowing it to release chatbots under either non-commercial and commercial licenses. It also releases the OpenAssistant Conversations (\texttt{oasst1}) dataset, which contains 66k conversations generated by human, accompanied with quality ratings. It also includes human preferences for the model responses, which enables RLHF training. After fine-tuning on this dataset, Open Assistant achieves a 48.3\% v.s. 51.7\% as compared to ChatGPT. As a high quality human-generated dataset free of GPT-synthesized content, \texttt{oasst1} is widely used in follow-up works.

\paragraph{StableVicuna}
    After the release of the \texttt{oasst1} dataset,  \citep{StableVicuna} proposes StableVicuna, ``the AI world's first open-source RLHF LLM chatbot''. It is fine-tuned on the Vicuna-13B model using a mix of the prompt-response datasets from Open Assistant, GPT4All, and Alpaca. The model is further optimized using RLHF with human preference data from Open Assistant, HH-RLHF \citep{hh-rlhf}, and SHP \citep{shp}. By the time StableVicuna is released, it outperforms other similarly sized open-source chatbots on a number of question-answering benchmarks.

\paragraph{Guanaco} 
    \citep{Dettmers2023QLoRA} introduces an efficient fine-tuning approach called QLoRA. As a by product, the chatbot Guanaco-65B fine-tuned on top of LLaMA achieves state-of-the-art results in human evaluation. It also releases the 7B/13B versions which are of a similar scale as previously mentioned small LMs. The fine-tuning dataset is a mix of \texttt{oasst1} \citep{OpenAssistant} and some other public datasets.

\subsection{Community trends and research directions}

In addition to trained models shown above, we would like to point out a few research trends around the topic of making small language models more efficient and performant. We discuss studies on accelerated training for large language models, performance improvement strategies, the scaling rules of large models, as well as the evaluation frameworks.

\paragraph{Acceleration and optimization}

\citet{Hewitt2023Backpack} propose Backpack, a new network architecture that takes all of performance, interpretability and control into consideration.
In Backpack, each word in a vocabulary is associated with multiple learned non-contextual sense vectors, and a word in a sequence is represented as a context-dependent, non-negative linear combination of its associated sense vectors.
The authors show that a 170M-parameter Backpack LM on OpenWebText has a comparable loss of a 124M parameter GPT-2 small, and,  Backpack sense vectors outperform word embeddings of a 6B-parameter Transformer LM on lexical similarity evaluations.

\citet{Liu2023Sophia} propose Sophia, Second-order Clipped Stochastic Optimization, an optimizer with light-weight estimate of the diagonal Hessian as the pre-conditioner to improve the popular, state-of-the-art optimizer Adam. 
Sophia attains half the number of steps, total compute, and wall-clock time compared with Adam with GPT-2 of sizes from 125M to 770M. The authors also prove theoretical properties of Sophia.

\citet{Lin2023AWQ} propose Activation-aware Weight Quantization (AWQ), "a hardware-friendly approach for LLM low-bit weight-only quantization", exploiting the observation that "protecting only 1\% of salient weights can greatly reduce quantization error".

\paragraph{Performance improvement}

\citet{Liu2023Goat} propose a fine-tuned LLaMA-based model Goat to outperform GPT-4 on arithmetic tasks, due to consistent tokenization of numbers by LLaMA. The authors decompose challenging tasks like multi-digit multiplication and division into learnable tasks and leverage basic arithmetic principles. The authors show that Goat-7B can be trained with LoRA on a 24GB VRAM GPU.

\citet{Patil2023Gorilla} propose a finetuned LLaMA-based model Gorilla to surpass GPT-4 on writing API calls.
With a document retriever, Gorilla adapts to document changes like user updates and version changes and mitigates hallucination. 
The author also introduce APIBench, a dataset including HuggingFace, TorchHub, and TensorHub APIs.

\paragraph{Study of the scaling law}

\citet{Eldan2023TinyStories} show that LMs with <10M parameters and one Transformer block can generate fluent and consistent stories of several paragraphs with close to perfect grammar. 

\citet{Gunasekar2023phi} introduce phi-1 and show good coding performance with 1.3B parameters and 7B training tokens, with a selection of “textbook quality” data.

\citet{Deshpande2023Scale} study downscaling effects with the shrunk language, showing the benefits of pre-training for models of 1.25M parameters and that compute-optimal models break the power law.
\citet{Mckenzie2023Inverse} provide 11 datasets  for empirical analysis of inverse scaling laws and discuss the important of data and objectives for training LMs.
\citet{Zhang2023Negation} propose NeQA, a dataset containing questions with negation and exhibit inverse scaling, U-shaped scaling, or positive scaling. Before this, the popular view follows scaling laws that the overall cross-entropy loss of an LM improves with the increased scale of model, dataset and compute for training~\citep{Kaplan2020Scaling}, and that the model and data should be scaled equally for compute-optimal training~\citep{Hoffmann2022Chinchilla}.

\paragraph{Evaluation for instruction-following LMs}

Fairly assessing the performance of instruction-following LMs poses a challenging task, given the extensive variety of tasks it must handle, including question answering, mathematics problem solving, coding and debugging, translation, and more. Furthermore, assessing the quality of chatbot responses is highly subjective in nature.  

Most works in Section \ref{sec:small_lm_gpt} and \ref{sec:small_lm_human} are evaluated by a few human evaluators on a small sample data. For instance, Alpaca \citep{Taori2023Alpaca} is evaluated by five students on around two hundred comparisons against \texttt{text-davinci-003}. Koala \citep{Geng2023Koala} is evaluated by 100+ people on 180 test queries. Open Assistant \citep{OpenAssistant} is evaluated using 7,042 manual comparisons on a sample of 22 prompts. 

On the other side, Vicuna \citep{Chiang2023Vicuna} employs GPT-4 as a proxy evaluator across 80 questions. This approach gains further support from Guanaco \citep{Dettmers2023QLoRA}, wherein both GPT-4 and humans are used to evaluate 953 user queries. The comparison demonstrate that GPT-4 evaluations serve as a ``cheap and reasonable'' substitute for human evaluation.

Evaluation of LMs in general, not just the instruction-following ones, continues to be a significant challenge and an active area of research. We delve deeper into this topic in Section \ref{sec:evaluation}.

\subsection{Evaluation}
\label{sec:evaluation}

Evaluation feedback is valuable for researchers and engineers to improve learning algorithms.
Evaluation and benchmarks for natural language processing, in particular, language models and  interactive applications, have been enjoying steady progress.
However, it is still challenging for research and development.


\citet{Burnell2023Evaluation} present guidelines for robust evaluation practices with more granular reporting, in particular, in-depth performance breakdowns beyond aggregate metrics and  instance-by-instance evaluation results.

\citet{Gehrmann2022Evaluation} survey obstacles in evaluation of test generation and propose to evaluate a model with multiple datasets via multiple metrics and document human evaluation well.
The authors propose the following best best practice \& implementation: make informed evaluation choices and document them, measure specific generation effects, analyze and address issues in the used dataset(s), evaluate in a comparable setting, run a well-documented human evaluation, produce robust human evaluation results, document results in model cards, and release model outputs and annotations.

\citet{Srivastava2022BIGBench} propose the Beyond the Imitation Game benchmark (BIG-bench) with more than 200 tasks.

\citet{Liang2022HELM} propose Holistic Evaluation of Language Models (HELM) to improve transparency of LMs,
with 1) a taxonomy of LM evaluation design space w.r.t. scenarios and metrics,
2) a broad coverage of 16 core scenarios with 7 metrics, i.e., accuracy, calibration, robustness, fairness, bias, toxicity, efficiency,
together with 7 targeted evaluations of skills and risks and 21 new scenarios,
and 3) evaluation of 30 existing models.

\citet{Lee2022HALIE} propose Human-AI Language-based Interaction Evaluation (HALIE) beyond non-interactive evaluation by
considering targets (full process and final output), perspectives (first-person and third-party), and criteria (preference and quality).

 Pythia~\citep{Biderman2023Pythia} is a suite of 16 LMs with sizes from 70M to 12B parameters and public access to checkpoints for each models to analyze the developments and evolutions of LMs over the course of training. 

\citet{Shumailov2023Curse} discuss the issue of model collapse due to training with generated data from LMs and show the importance of genuine human data for LMs.

\section{Applying “Mini-Giants” to real-world}
\label{section:application}
"Mini-giants" are uniquely positioned to solve two important issues unaddressed by larger language models: privacy protection and local computation. We examine the application of these smaller models in real-world scenarios, using the therapeutic chatbot Woebot as an example. Cognitive Based Therapy (CBT) took several years from being popular in Woebot, to become closer to clinical ready.

Before delving into the discussion, let's clarify the definition of small language models. Recall that by today's standard, small LMs are the models with parameter sizes of around 10B or lower and with performance comparable or better than ChatGPT / GPT-4. However, this is a definition based on today’s technology capabilities. With the development of hardware and other optimization softwares, there will definitely be “mini-giants” with  much more network parameters in the future. Therefore, to future-proof our discussion on applications, we use a more extensible definition for a “mini-giant”: a language model which can be trained/modified/used with affordable resources, like with a single GPU and an open source developer today.

Compared to their larger counterparts, “Mini-giants” offer two advantages: privacy protection and computation efficiency. Users wishing to utilize language models have two primary choices. They can either utilize APIs provided by organizations like OpenAI, or build their own "mini-giants". If they choose the former, it is expected that their proprietary data will go through third party's servers and be logged, which would be unacceptable to sensitive industries such as financial or health care institutes. On the other hand, "Mini-giants" permit centralized user data storage, potentially on a single GPU. For example, “Alpaca-Lora” can run locally on affordable hardware like a Raspberry Pi.
In terms of computation efficiency, in industries like autonomous driving, high network latency may occur when connecting to remote data centers. Hence, it's crucial that the language model can function independently.

To demonstrate "mini-giants" advantages, we examine Cognitive Based Therapy (CBT), an effective technique for treating clinical depression. Moving CBT from casual to clinical use is a demanding process, involving extensive clinical trials. Woebot, an AI chatbot, incorporates CBT into daily use, providing around-the-clock mental health support and anxiety reduction. The company Woebot was founded by Alison Darcy, a psychology student who worked as a software engineer, and then joined Stanford as a postdoctoral researcher in clinical psychology in 2017. Since its establishment, it received endorsement from AI pioneers such as Andrew Ng, who became one of the board of directors in 2017. The chatbot is a popular App with 4.7 rating out of 5, and more than 5,900 reviews in July 2023, and exchanges millions of messages with users every week in 2021~\citep{WoebotMichaelEversAI}.

However, despite great user reviews, it took more than two years for the company to go through the clinical trials process and get closer to being endorsed by mental health doctors. Woebot first posted their clinical trials recruitment notice on ClinicalTrials.gov in 2019, and designed a process to recruit 101 participants to evaluate whether this chatbot can help in alcohol use disorders etc. It took around 5 months to complete the study in 2020, and the results were first posted in Aug 2022. \citep{WoebotClinical}. In 2023, Woebot announced the enrollment of the first patient in a pivotal clinical trial to evaluate if it can help women with postpartum depression \citep{WoebotWB001}. Their paper published in Expert Review of Medical Devices \citep{WoebotAnatomyWB001} documented the clinical trial process.

The reader might ask why it takes such a complicated experimentation process to adopt a new technology in clinical trials and go through the U.S. Food \& Drug Administration (FDA) process. The answer is simple. If your families and friends are going to go to a doctor and look for some mental health help, what evidence would you need to decice a chatbot is as trust-worthy as a doctor?

In short, “mini-giants” a.k.a. “small” language models had some unique advantages in privacy protection and computation efficiency. However, their successful integration into specific domains like healthcare requires adherence to industry standards, a frequently long process involving more than just technological considerations.

\section{Discussion and outlook}

As the capability of large foundation models and AI becomes increasingly well-known to the general public, the demand for AI democracy becomes an issue of societal fairness and equity. In our opinion, the open source community and "small" language models mark one step towards facilitating AI democracy, making it easier for everyone to control, adapt, interpret and afford the power of AI. 


\begin{itemize}
  
  \item \textbf{Adaptability}: For the open source communities including Kaggle, the ability to innovate comes from the capability to use the model in ways that are best suited to domain specific scenarios. Prompt engineering alone is not enough. Thanks to methods mentioned in Section \ref{sec:methods}, fine-tuning even complex model architectures can mostly be achieved on a single or a few GPUs. Without this, the role of ML researchers without an unimaginable amount of resources risks being diminished to prompt engineers.
  
  \item \textbf{Controllability}: Being able to choose where to run the model, what data is seen by the model, and what model outputs are used relies heavily on the model being easy enough to run on local infrastructure, and model components are transparent and interpretable. Section \ref{sec:survey} listed a wide range of options to select from for research and/or business use, which leverage the power of large foundation models and at the same time keeps data local. Moreover, with smaller models, users will have a better chance tuning it with instruction following strategies, to further reduce mis-information and ensure the compliance requirements for model outputs. This increases the chance of successful AI application in compliance-demanding domains.
  
  \item \textbf{Affordability}: Having access to smaller models and cheaper training / fine-tuning options is the only way that privacy-sensitive industries and applications can avoid the trade-off between giving up the right of autonomous data governance, and squandering unreasonable amounts of funds on training gigantic models in-house. As mentioned in Section \ref{section:application}, the affordable option to build domain-specific "small" language models enables industries like finance and healthcare to leverage AI without risking leaking sensitive data to unwarranted third parties. In this sense, lowered costs brought about by these "small" models can prevent the privilege of using AI from falling into the hands of a few exclusive entities.
\end{itemize}

To sum up, being users of new achievements like GPT-4 is great. Being builders and/or owners of innovations is even better. As technology optimists, the authors believe that it is only through the ability to understand and leverage AI that the society as a whole can mitigate the potential AI risks. With a well designed paradigm, the open source community ans small language models can increase the chance for all to benefit from, and to contribute to, the power of AI.



\section*{Acknowledgments}

The authors would like to thank Yiyao Liu and Qibin Chen for offering constructive feedback and valuable insights. 
The authors used ChatGPT \citep{ChatGPT} to edit several sentences in the essay with the following prompt: \emph{Revise to more concise, formal, and fluent, following the style of an academic research paper: [Insert sentence]}.

\bibliography{main}
\bibliographystyle{acl_natbib}

\end{document}